\documentclass[10pt,twocolumn,letterpaper]{article}

\usepackage{graphicx}
\usepackage{amsmath}
\usepackage{amssymb}
\usepackage{booktabs}
\usepackage[pagebackref,breaklinks,colorlinks]{hyperref}
\usepackage{pgfplots}
\usepackage{hyperref}
\usepackage{tabularx}
\usepackage{adjustbox}
\usepackage{float}
\usepackage{dblfloatfix}
\usepackage{caption}
\RequirePackage[font=footnotesize,skip=3pt,subrefformat=parens]{subcaption}

\usepackage[capitalize]{cleveref}
\crefname{section}{Sec.}{Secs.}
\Crefname{section}{Section}{Sections}
\Crefname{table}{Table}{Tables}
\crefname{table}{Tab.}{Tabs.}

\def\etal{\emph{et al.}}

\begin{document}

\title{Anisotropic multiresolution analyses for deepfake detection}

\author{\\
Wei Huang\\
Istituto Eulero\\
Universit\`a della Svizzera italiana\\
{\tt\small }
\and
\\
Michelangelo Valsecchi\\
Istituto Eulero\\
Universit\`a della Svizzera italiana\\
{\tt\small }
\and
\\
Michael Multerer\\
Istituto Eulero\\
Universit\`a della Svizzera italiana\\
{\tt\small}
}
\maketitle

%%%%%%%%% ABSTRACT
\begin{abstract}
Generative Adversarial Networks (GANs) have paved the path towards
entirely new media generation capabilities at the forefront of
image, video, and audio synthesis. However, they can also be misused
and abused to fabricate elaborate lies, capable of stirring up the public debate.
The threat posed by GANs has sparked the need to discern between genuine content
and fabricated one. Previous studies have tackled this task by using classical
machine learning techniques, such as k-nearest neighbours and eigenfaces, which unfortunately
did not prove very effective. Subsequent methods have focused on leveraging on
frequency decompositions, i.e., discrete cosine transform, wavelets, and wavelet packets,
to preprocess the input features for classifiers. However, existing approaches only rely
on isotropic transformations. We argue that, since GANs primarily utilize isotropic
convolutions to generate their output, they leave clear traces, their fingerprint,
in the coefficient distribution on sub-bands extracted by anisotropic transformations.
We employ the fully separable wavelet transform and multiwavelets to obtain the anisotropic
features to feed to standard CNN classifiers. Lastly, we find the fully separable transform
capable of improving the state-of-the-art.
\end{abstract}

%%%%%%%%% BODY TEXT
\section{Introduction}
Generative Adversarial Networks (GANs) have become a thriving topic in recent years
after the initial work by Goodfellow \etal in \cite{goodfellow2014generative}.
Since then, GANs have quickly become a popular and rapidly changing field due to their
ability to learn high-dimensional complex real image distributions. As a result,
numerous GAN variants have emerged, like CramerGAN (\cite{bellemare2017cramer}),
MMDGAN (\cite{li2017mmd}), ProGAN (\cite{gao2019progan}), SN-DCGAN (\cite{miyato2018spectral}),
and the state-of-the-art StyleGAN, StyleGAN2, and StyleGAN3 (\cite{karras2019style,karras2020analyzing,karras2021alias}). Among various primary
applications of GANs is fake image and video generation, e.g., Deepfakes (\cite{deepfakes}),
FaceApp (\cite{faceapp}), and ZAO (\cite{zao}). In particular, Deepfakes is the first successful
project taking advantages of deep learning, which was started in 2017 on Reddit by an account
with the same name. Since then, deepfakes are regarded as falsified images and videos created
by deep learning algorithms, see \cite{sencar2022multimedia}. A major source of motivation for
investigation into the automatic deepfake detection is the visual indistinguishability between
fake images created by GANs and real ones. Moreover, the abuse of fake images potentially
pose threats to personal and national security. Therefore, research on deepfake detection has
become increasingly important with the rapid iteration of GANs. 

There are two kinds of tasks in the detection of GAN-generated images. The easiest is identifying
an image as real or fake. The harder one consists of attributing fake images to the corresponding
GAN that generated them. In this paper, we mainly focus on the attribution task. Both tasks involve
extracting features from images and feeding them to classifiers. For the classifiers, there are
approaches based on traditional machine learning methods, which are relatively simple,
but often reach relatively bad results, see \cite{fridrich2012rich,khan2019benchmark}. 
Approaches based on deep learning, especially convolutional neural networks (CNN), have proven
powerful and are employed in many recent papers, see \cite{rossler2019faceforensics++, yu2019attributing,Wang_2020_CVPR,liu2020global, yu2020responsible,frank2020leveraging, wolter2021wavelet}. 
For feature extraction, the simplest method is just using raw pixels as input. The results are, however,
not of high accuracy and the classifiers fed with raw pixels are not robust under common perturbations, see \cite{liu2020global,frank2020leveraging}. Therefore, it is necessary to develop methods to better extract
features. One stream is the learning-based method by Yu \etal in \cite{yu2019attributing,yu2021artificial, yu2020responsible}, which found unique fingerprints of each GAN. Another stream is based on the mismatches
between real and fake in the frequency domain, see \cite{zhang2019detecting,durall2019unmasking,frank2020leveraging,durall2020watch,liu2020global,wolter2021wavelet}. Specifically, multiresolution methods, e.g., the wavelet packet transform, have recently been employed for
deepfake detection, see Wolter \etal in \cite{wolter2021wavelet}. Their work demonstrates the capabilities
of multiresolution analyses for the task at hand and marks the starting point for our considerations.
In contrast to the isotropic transformations considered there, we focus on anisotropic transformations,
i.e., the fully separable wavelet transform (\cite{velisavljevic2006directionlets}) and samplets (\cite{harbrecht2022samplets}), which are a particular variant of multiwavelets. 

Because the generators in all GAN architectures synthesize images in high resolution
from low resolution images using deconvolution layers with square sliding windows,
it is highly likely for the anisotropic multi wavelet transforms of fake images to leave
artifacts on anisotropic sub-bands. In this paper, we show that features from anisotropic (multi-)wavelet
transforms are promising descriptors of images. This is due to remarkable mismatches between the anisotropic
multiwavelet transforms of real and fake images, see Figure \ref{fig:fp}. To evaluate the anisotropic features,
we set up a lightweight multi-class CNN classifier as in \cite{frank2020leveraging, wolter2021wavelet}
and compare our results on the datasets consisting of authentic images from one of the three commonly
used image datasets: Large-scale Celeb Faces Attributes (CelebA \cite{celeba}), 
LSUN bedrooms (\cite{lsun}), and Flickr-Faces-HQ (FFHQ \cite{karras2019style}),
and synthesized images generated by CramerGAN, MMDGAN, ProGAN, and SN-DCGAN on the CelebA and LSUN bedroom,
or the StyleGANs on the FFHQ. Finally, as in \cite{frank2020leveraging,wolter2021wavelet},
we test the sensitivity to the number of training samples and the robustness under the four common perturbations: Gaussian blurring, image crop, JPEG based compression, and addition of Gaussian noise.

\section{Related work}
\noindent\textbf{Deepfake detection:}
A comprehensive statistical studying of natural images shows that regularities always exist
in natural images due to the strong correlations among pixels, see \cite{lyu2013natural}.
However, such regularity does not exist in synthesized images. Besides, it is well-known that
checkerboard artifacts exist in CNNs-generated images due to downsampling and upsampling layers,
see examples in \cite{odena2016deconvolution, azulay2018deep}. The artifacts make identification
of deepfakes possible. In \cite{marra2018detection, rossler2019faceforensics++,Wang_2020_CVPR},
the authors show that GAN-generated fake images can be detected using CNNs
fed by conventional image foresics features, i.e., raw pixels. In order to improve the accuracy
and generalization of classifier, several methods are proposed to  address the problem of finding more discriminative features instead of raw pixels. Several non-learnable features are proposed,
for example hand-crafted cooccurrence features in \cite{nataraj2019detecting}, color cues in \cite{mccloskey2018detecting}, layer-wise neuron behavior in \cite{wang2019fakespotter},
and global texture in \cite{liu2020global}. In \cite{yu2019attributing},
Yu \etal discover the possibility of uniquely fingerprinting each GAN model and
characterize the corresponding output during the training procedure.
With this technique, responsible GAN developers could fingerprint their models and keep track
of abnormal usage of their releases. In the follow-up paper (\cite{yu2020responsible}),
Yu \etal scale up the GAN fingerprinting mechanism. However, in \cite{neves2020ganprintr},
Neves \etal propose GANprintR to remove the fingerprints of GANs,
which renders this identification method useless.

\noindent\textbf{Frequency artifacts:}
It is found that artifacts are more visible in the frequency domain.
State of the art results are achieved using features in the frequency domain, e.g.,
the coefficients of the discrete cosine transform (\cite{zhang2019detecting,durall2019unmasking,frank2020leveraging,durall2020watch,liu2020global}) and the coefficients of the isotropic wavelet packet transform (\cite{wolter2021wavelet}).
In \cite{frank2020leveraging}, Frank \etal found that the grid-like patterns in the frequency domain
stem from the upsampling layers. Even though ProGAN and StyleGANs are equipped with improved upsampling
methods, artifacts still exit in their frequency domain. Combination of the features in frequency domain and lightweight convolutional neural networks can outperform the complex heavyweight convolution neural networks
using features based on the pixel values of images. In \cite{wolter2021wavelet}, features based on wave-packets
are used, which outperforms all the other state-of-the-art methods with comparable lightweight CNN classifiers.
The success of the isotropic wave-packets inspired us to further investigate this direction and to
also take into account anisotropic multiresolution analyses,
to extract more distinguishable features for the deepfake detection.

\section{Proposed Method}
\subsection{Motivation}
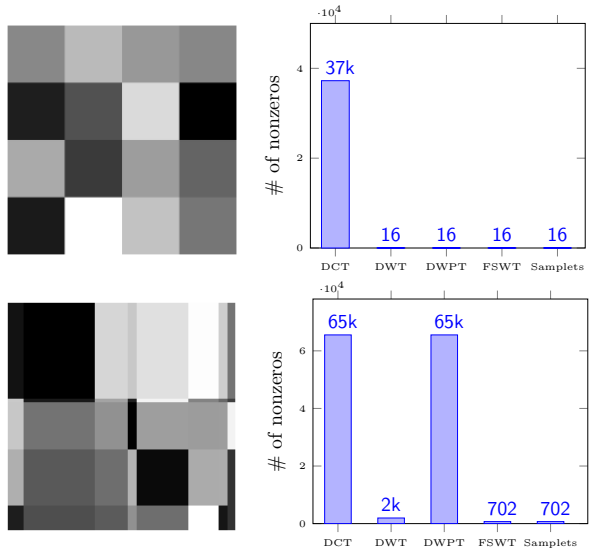
\begin{figure}[ht]
\begin{subfigure}{.4\columnwidth}
  \centering
  \includegraphics[width=\linewidth]{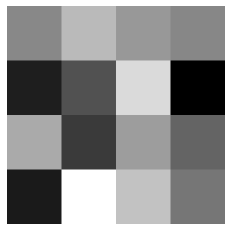}
\end{subfigure}%
\begin{adjustbox}{width=0.6\columnwidth}
\begin{subfigure}{.8\columnwidth}
\centering
\begin{tikzpicture}
\pgfplotsset{compat=1.16, every tick label/.append style={font=\tiny}, every node near coord/.style={font=\Tiny}}
\begin{axis}[ybar,bar width=0.5, % NEW BIT
xtick={0,1,2,3,4},  % NEW BIT
xticklabels={DCT,DWT,DWPT,FSWT,Samplets},
ylabel={\# of nonzeros},
ymin=0,
ymax=50000,
xmax=4.5,
width=\textwidth, % NEW BIT
nodes near coords style={font=\sffamily,align=center,text width=1em},
nodes near coords=\pgfmathsetmacro{\mystring}{{"37k","16","16","16","16"}[\coordindex]}\mystring,
nodes near coords align={vertical},
]
\addplot coordinates
% Transfer
{(0,37249) (1,16) (2,16) (3,16) (4,16)};
\end{axis}
\end{tikzpicture}
\end{subfigure}
\end{adjustbox}
\begin{subfigure}{.4\columnwidth}
  \centering
  \includegraphics[width=\linewidth]{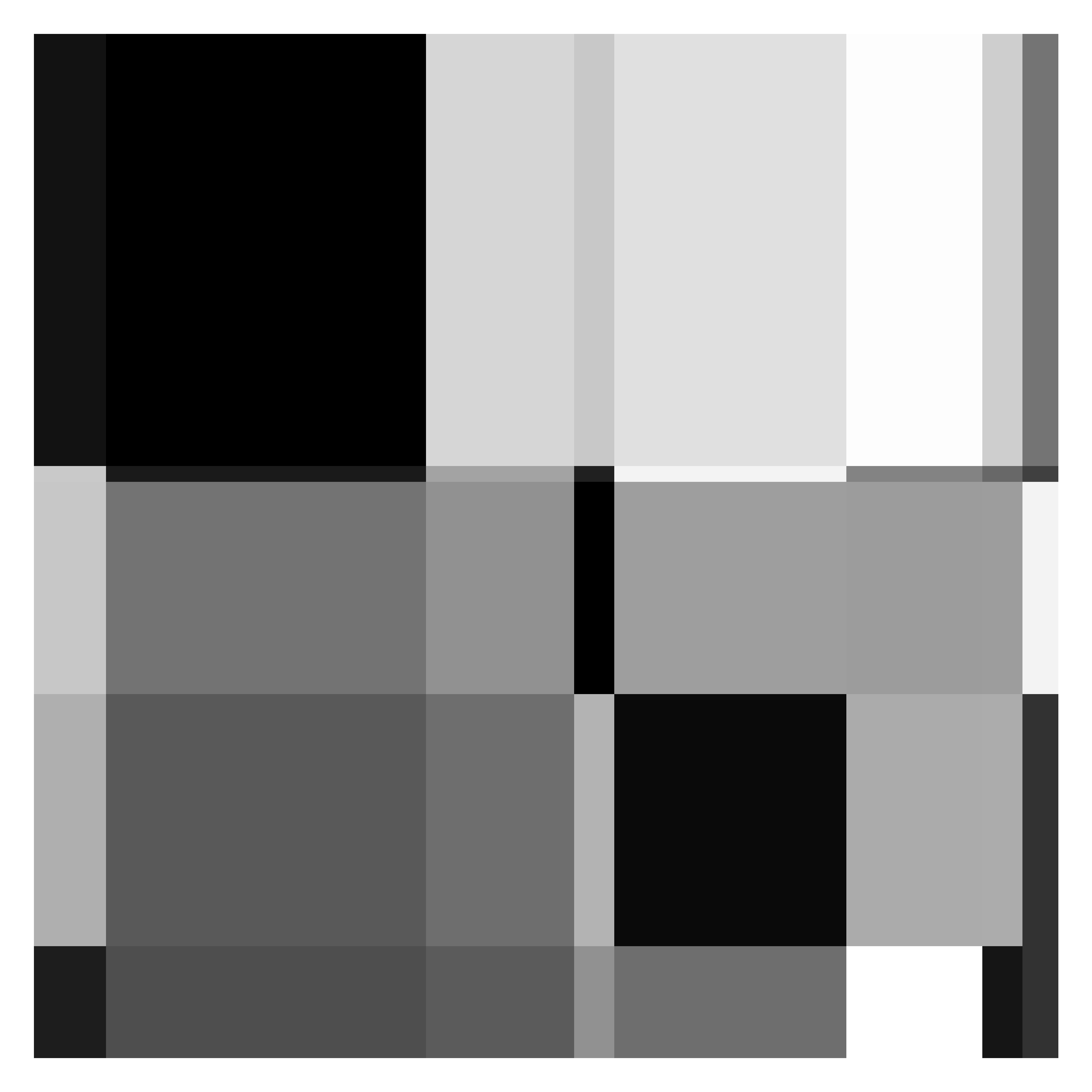}
\end{subfigure}%
\begin{adjustbox}{width=0.6\columnwidth}
\begin{subfigure}{.8\columnwidth}
\centering
\begin{tikzpicture}
    \pgfplotsset{compat=1.16, every tick label/.append style={font=\tiny}, every node near coord/.style={font=\tiny}}
    \begin{axis}[ybar,bar width=0.5, % NEW BIT
    xtick={0,1,2,3,4},  % NEW BIT
    xticklabels={DCT,DWT,DWPT,FSWT,Samplets},
    ylabel={\# of nonzeros},
    ymin=0,
    ymax=78000,
    xmax=4.7,
    width=\textwidth, % NEW BIT
    nodes near coords style={font=\sffamily,align=center,text width=1em},
    nodes near coords=\pgfmathsetmacro{\mystring}{{"65k","2k","65k","702","702"}[\coordindex]}\mystring,
    nodes near coords align={vertical},
    ]
    \addplot coordinates
    % Transfer
    {(0,65536) (1,1949) (2,65536) (3,702) (4,702)};
    \end{axis}
\end{tikzpicture}
\end{subfigure}
\end{adjustbox}
\caption{The top left part shows a \(256\times256\) blockwise constant grayscale image
with 16 squares of equal size (black stands for 0, and white stands for 255). The bottom left
part shows a \(256\times256\) blockwise constant grayscale image with 40 rectangles
(black stands for 0, and white stands for 255). The right part shows the numbers of
nonzero coefficients under different transforms of the isotropic and anisotropic patterns
respectively. Herein, we consider the wavelets and samplets with vanishing moments up to order 1.}
\label{fig:iso_and_aniso_patterns}
\end{figure}

Images are often composed of two types of regions:
mostly monochromatic patches, usually backgrounds, and areas with sharp color
gradients, found in correspondence with borders that separate different objects.
This construction is similar to a square wave in 1D, which is notoriously difficult
to approximate with only cosine functions like the discrete cosine transform (DCT) does.
This fact is known as the Gibbs phenomenon, see \cite{gibbs1898fourier}. Similar to a square
wave in 1D, images can be considered as piecewise constant functions in 2D, which makes using
DCT methods challenging as their supports are not localized in space but only in frequency.
This results in redundant representations of images in the frequency domain. One solution,
proposed in \cite{wolter2021wavelet}, is to decompose an image into frequencies while also
maintaining spatial information is using wavelets, which are localized in both domains and
are thus less susceptible to discontinuities. In order to manifest the efficient representation
of images using wavelets, we consider an isotropic pattern with discontinuities on the boundaries
of square blocks, and a anisotropic pattern with discontinuities on the boundaries of rectangular
blocks, see Figure \ref{fig:iso_and_aniso_patterns}. We then compute the DCT and four different
kinds of wavelet transforms, i.e., the discrete wavelet transform (DWT), the discrete wavelet packet
transform (DWPT), the fully separable wavelet transform (FSWT), and the samplet transform.
From the bar plot in Figure \ref{fig:iso_and_aniso_patterns}, all wavelet transforms 
overcome the Gibbs phenomenon, in contrast to the DCT. However, anisotropic wavelet
transforms, i.e., FSWT and samplets, perform much better than isotropic DWT and DWPT in the task of finding
efficient representations for anisotropic patterns which commonly exist in real images.

The previous works \cite{zhang2019detecting,frank2020leveraging} have already analyzed
the effectiveness of using the frequency domain instead of the direct pixel representation
when detecting deepfakes. Moreover, the method in \cite{wolter2021wavelet} has improved
the state-of-the-art result using the isotropic wavelet, i.e., the DWPT. However, they usually
result in redundant representations of images. Moreover, they rely on only isotropic decompositions.
We are convinced that anisotropic transforms can add a new aspect to the challenge at hand.
The intuition behind this reasoning comes from the fact that GAN architectures typically only use
isotropic convolutions (square sliding windows) to synthesize new samples, thus being unaware of the
fingerprint they are leaving in the hidden anisotropic coefficients' distribution of the image. 

We focus on two technologies that allow us to expose these fingerprints and obtain
a spatio-frequency representation of the source image: the fully separable wavelet transform and samplets,
which are a particular variant of multiwavelets.

\subsection{Preliminaries}
\subsubsection{Isotropic wavelets}
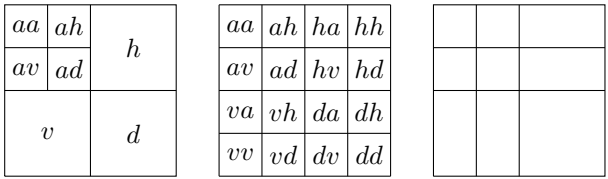
\begin{figure}[ht]
\centering
\begin{tikzpicture}[scale=0.57]
\draw[black, thin] (0,0) -- (4,0) -- (4,4) -- (0, 4) -- 
(0,0);
\draw[black, thin] (2,0) -- (2,4);
\draw[black, thin] (1,2) -- (1,4);
\draw[black, thin] (0,2) -- (4,2);
\draw[black, thin] (0,3) -- (2,3);
\node (v) at (1,1) {$v$};
\node (d) at (3,1) {$d$};
\node (h) at (3,3) {$h$};
\node (aa) at (0.5,3.5) {$aa$};
\node (av) at (0.5,2.5) {$av$};
\node (ad) at (1.5,2.5) {$ad$};
\node (ah) at (1.5,3.5) {$ah$};
\foreach \x in  {5,...,9} {%
    \draw[black, thin] (\x,0) -- (\x,4);
}
\foreach \y in  {0,...,4} {%
    \draw[black, thin] (5,\y) -- (9,\y);
}
\node at (5.5,3.5) {$aa$};
\node at (5.5,2.5) {$av$};
\node at (6.5,2.5) {$ad$};
\node at (6.5,3.5) {$ah$};

\node at (7.5,3.5) {$ha$};
\node at (7.5,2.5) {$hv$};
\node at (8.5,2.5) {$hd$};
\node at (8.5,3.5) {$hh$};

\node at (5.5,1.5) {$va$};
\node at (5.5,0.5) {$vv$};
\node at (6.5,0.5) {$vd$};
\node at (6.5,1.5) {$vh$};

\node at (7.5,1.5) {$da$};
\node at (7.5,0.5) {$dv$};
\node at (8.5,0.5) {$dd$};
\node at (8.5,1.5) {$dh$};

\draw[black, thin] (10,0) -- (14,0) -- (14,4) -- (10, 4) -- 
(10,0);
\draw[black, thin] (10,2) -- (14,2);
\draw[black, thin] (10,3) -- (14,3);
\draw[black, thin] (12,0) -- (12,4);
\draw[black, thin] (11,0) -- (11,4);
\end{tikzpicture}%
    \caption{Leftmost: the quad tree structure of the wavelet decomposition of an image.
     Middle: the complete tree structure of a wavelet packet.
     Rightmost: tree structure of a fully separable wavelet decomposition.}
    \label{fig:wvlt_structures}
\end{figure}

\noindent\textbf{Discrete wavelet transform:} Wavelets are localized waves that have a nonzero
value around a certain point but then collapse to zero when moving further away. Examples for such
wavelets are Haar- and Daubechies- wavelets, see \cite{daubechies1992}. The main idea behind
wavelet-based analyses is the decomposition of a signal with respect to a hierarchical scales.
The smaller the scales, the higher the corresponding frequency. Unlike the DCT, wavelets are localized
in the spatial domain as well, due to their hierarchical nature. The fast wavelet transform (FWT),
see \cite{beylkin1991fast},
is an algorithm commonly used to apply a discrete wavelet transform onto an $n\times n$ image
and decompose it into approximation coefficients using convolutions of filter banks and downsampling
operators, with a computational cost in the order of $\mathrm{O}(n^2)$, i.e., linear in the number of pixels.
The results of one decomposition step of the FWT are four sets of coefficients usually referred to as
$a$, $h$, $v$, $d$, which stand for approximation, horizontal, vertical, and diagonal coefficients.
To produce a decomposition up to level $l$, the $a$ coefficient of level $l-1$ is further decomposed
into four components: $aa$, $ah$, $av$, $ad$, giving rise to the structure in the leftmost side of
Figure \ref{fig:wvlt_structures}. This approach is also known as Mallat decomposition (\cite{mallat_dec})
and amounts to a 2D isotropic wavelet transform. All frequency-based methods like Fourier transformations
and wavelets are usually tailored towards unbounded domains, while images are instead bounded.
In the case of the Haar wavelet, the transition from unbounded to bounded is not a problem and
no boundary modifications are required to obtain an orthonormal basis.
However, higher-order wavelets require modifications at the boundary,
which introduces the need for padding, usually accomplished either by appending zeros to a dimension
or by repeating the data or reflecting it. To avoid changing the size of images, 
the Gram-Schmidt boundary filter is proposed in \cite{boundary_filters}, which replaces the wavelet
filters at the boundary with special shorter filters that preserve both the shape and the
reconstruction property of the wavelet transform. 

\noindent\textbf{Wavelet packets:} Following the assumption that information is mostly contained
in the low-frequency approximation coefficient $a$, the Mallat approach leaves the high frequency
terms $h$, $v$, and $d$ untouched when moving up one level in the decomposition.
Previous work (\cite{fft_lower_freq}) has instead shown that,
especially when dealing with compressed and lower resolution images, information in the higher
frequencies is usually affected more by the reduction processes than that in the lower part of the
frequency spectrum. Therefore, the remaining three feature maps are likely to also contain
relevant information for deepfake detection. Motivated by this reasoning,
the authors of \cite{wolter2021wavelet} employed wavelet packets as input features for their
classifiers. A single level decomposition of a wavelet packet is essentially the same as that
of the FWT, but when decomposing up to level $l$, also the $h$, $v$, and $d$ of level $l-1$ are
further decomposed into $(ha,hh,hv,hd)$, $(va,vh,vv,vd)$ and $(da,dh,dv,dd)$, at the total cost of
$\mathrm{O}(n^2\log n)$ operations. The resulting quad-tree-like structure is depicted in the middle
of Figure \ref{fig:wvlt_structures} for a complete decomposition. In \cite{wolter2021wavelet},
the authors used a Gram-Schmidt boundary filter such that coefficients of any image using the
wavelet packet transform remain the same size as the transformed image. Afterwards, 
all packets are stacked together along the channel direction before feeding the CNN. 

\subsubsection{Anisotropic wavelets}
\noindent\textbf{Fully Separable Wavelet Transform:} Previous work has focused on the
isotropic discrete wavelet transform and its derivatives to extract features. For images,
the isotropic decomposition from level 1 up to level $l$ consists of transforming each axis
once for each level. As we have seen with the FWT and wavelet packets, first the $x$ axis is
decomposed, then the $y$ axis, and only after the algorithm moves to the next level.
Instead, the fully separable decomposition first decomposes completely one axis up to level $l$ and
then decomposes the obtained features on the other axis
(see the rightmost of Figure \ref{fig:wvlt_structures} for a sketch of the resulting pattern).
This different approach gives rise to a more anisotropy-friendly feature extraction while only
increasing the computational cost by a factor of two compared to the Mallat decomposition.
Similarly to wavelet packets, a suitable boundary modification, e.g., by padding,
is necessary for the fully separable wavelet transform with more than one vanishing moment.
In this paper, we consider both the reflect-padding method and the 
boundary filter with QR orthogonalization. 

\noindent\textbf{Samplets:}
First presented in \cite{harbrecht2022samplets}, samplets are a generalization of
multiwavelets that produce a multiresolution analysis of a given signal in terms of
discrete signed measures. Samplets can be constructed with an arbitrary number $m$ of
vanishing moments for arbitrary data sets. For structured data, such as images, 
lowest order samplets (m=1) correspond to Haar wavelets. 
The construction of the basis and the fast samplet transform can both be performed in
linear time, i.e., the cost of transforming an $n\times n$
input image is \(\mathcal{O}(n^2)\).
Different from the FSWT, samplets are a data-centric approach and
can accommodate any data dimensions without the need
for padding or special boundary treatment. In particular,
they do not suffer from discontinuities at
the boundaries and do not introduce artifacts in the decomposition,
which allows using samplets directly without modifications
and without increasing their computational cost.
 
\subsection{Proof of concept}
\begin{figure*}[ht]
	\centering
	\includegraphics[width=\linewidth]{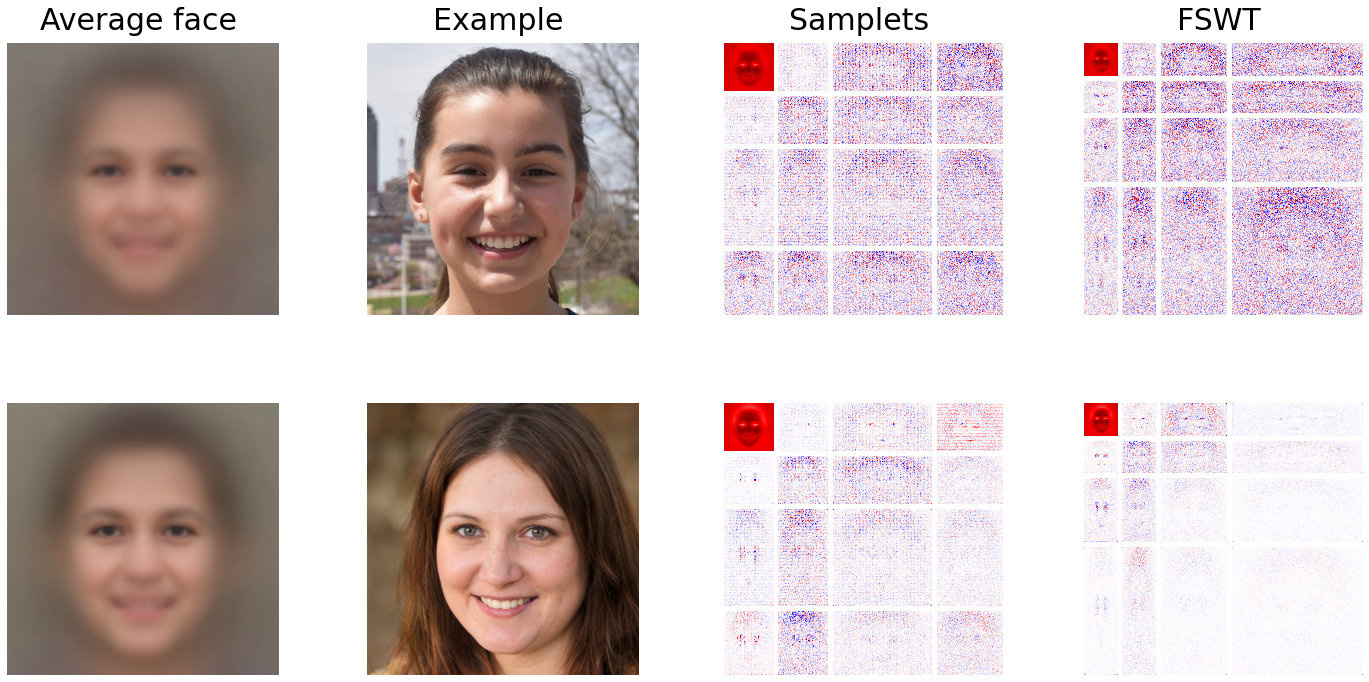}
	\caption{A row by row comparison of real and synthesized images on sub-bands. The first row shows
	the average of 1000 real images from the FFHQ dataset. The second column shows the
	corresponding block-wise normalized samplet coefficients, while the last column shows the
	block-wise normalized coefficients of the FSWT. 
	Herein, we employed samplets with vanishing moments up to order $3$ and level \(l=3\),
	and the FSWT with the Daubechies 3 orthogonal wavelet, reflect-padding method, and \(l=3\).
	The second row shows the same information for a dataset of the same size generated by
	the StyleGAN model taken from \href{thispersondoesnotexist.com}{thispersondoesnotexist.com}. In the
	 color map, white stands for zeros, red for positive numbers, and blue for negative numbers.}
	\label{fig:fp}
\end{figure*}

\begin{figure}[ht]
	\begin{subfigure}{.5\linewidth}
		\centering
		\includegraphics[trim={0.0cm 12.0cm 12.0cm 0.0cm},clip,width=\linewidth]{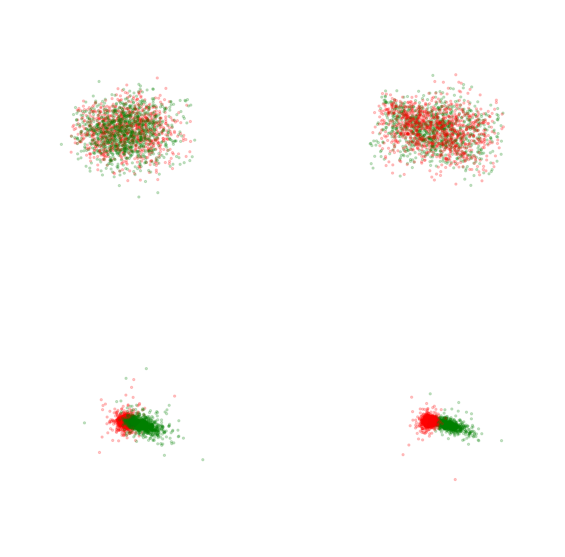}
		\caption{Raw pixels}
	\end{subfigure}%
	\begin{subfigure}{.5\linewidth}
		\centering
		\includegraphics[trim={12.0cm 12.0cm 0.0cm 0.0cm},clip,width=\linewidth]{figs/ffhq_fs_pca.png}
		\caption{Low frequency coefficients}
	\end{subfigure}
	\begin{subfigure}{.5\linewidth}
		\centering
		\includegraphics[trim={0.0cm 2.5cm 12.0cm 12.5cm},clip,width=\linewidth]{figs/ffhq_fs_pca.png}
		\caption{Anisotropic coefficients}
	\end{subfigure}%
	\begin{subfigure}{.5\linewidth}
		\centering
		\includegraphics[trim={12.0cm 2.5cm 0.0cm 12.5cm},clip,width=\linewidth]{figs/ffhq_fs_pca.png}
		\caption{High frequency coefficients}
	\end{subfigure}
	
	\caption{The 3D scatter plots of 2000 samples consisting of 1k real and 1k fake images visualize the first three principle components of the raw pixels and the three different parts of the FSWT coefficients.The red nodes represent real images, and the green ones represent fake images.}
	\label{fig:pca-ffhq}
\end{figure}
To illustrate how features are extracted by anisotropic multiresolution analyses,
we visualize in Figure \ref{fig:fp} the samplet- and FSTW- coefficients 
for the average of 1000 real images from the FFHQ (top row) and for
1000 images by the StyleGAN models from the photorealistic face generation website
\href{thispersondoesnotexist.com}{thispersondoesnotexist.com} (bottom row).
As can be seen, the average images from the real sample and the one generated by
StyleGAN in the first column are very similar. In particular, it is very difficult to tell 
which of the exemplary images in the second column is synthesized and which one is real.
According to the experiments in \cite{liu2020global}, human performance in classification
only leads to an accuracy of about 63.9\% in the FFHQ vs StyleGAN case, which is only slightly
higher than the probability of randomly guessing. The two anisotropic transforms lead
to the sparse patterns on the right side of the figure. The patterns for the real images
and the synthesized images are clearly distinguishable, especially by examining the
high and anisotropic frequency parts. These observations suggest that
samplets and FSWT are suitable for assisting classifiers in discerning the origin of a
given image.

To further illustrate the idea of using features extracted by the two proposed anisotropic
transforms, we visualize the first three principal components of the raw pixels and
different parts of the FSWT coefficients in a principal component analysis (PCA).
The samples for FFHQ and StyleGAN2 consist of 1000 images, each, with a resolution
of \(128\times 128\). Panel (a) in Figure \ref{fig:pca-ffhq} shows the principal
components of the raw pixels, which are not distinguishable. The same accounts for
the low frequency contributions depicted in panel (b). In turn,
the principal components of real and synthesized images become linearly separable
in the anisotropic parts and high frequency parts, depicted in panels (c) and (d).

The scales of the wavelet transform coefficients decrease exponentially with respect to
their underlying level \(l\) if the underlying signal exhibits sufficient smoothness.
However, the importance of the high frequency coefficients at high levels is large.
Therefore, we employ a block-wise normalization, where we divide coefficients by
the maximum absolute value on the corresponding block. This way, we can bring each
block to the same range \((-1,1)\). Moreover, we keep the zero coefficients untouched
to maintain the sparsity pattern. In the following experiments, this block-wise
normalization is always applied before feeding the data into the CNN classifier. 

\section{Experiments}
All experiments in this paper are performed on a compute server with
two Intel Xeon (E5-2650 v3 @ 2.30GHz),
one NVIDIA A100-PCIe-40GB, two NVIDIA GeForce GTX 1080 Ti (Titan 11GB GDDR5X),
and one NVIDIA GeForce GTX 1080 (Founders Edition 8GB GDDR5X).
\subsection{Source Identification}
\noindent\textbf{Experimental Setup}
The experiments are conducted on three datasets: CelebA, LSUN bedroom, and FFHQ.
The datasets CelebA and LSUN bedroom are generated by the pre-trained GAN models
from \cite{yu2019attributing}, and contain 150k resized real images of resolution
\(128\times128\) and 150k fake images in the same size for each model. 
The models under consideration are CramerGAN, MMDGAN, ProGAN, and SN-DCGAN.
Thus, the total dataset has a size of 750k.
It is then partitioned into training, validation, and test datasets consisting of
500k, 100k, and 150k images respectively.
On the other hand, the FFHQ dataset contains 70k real images resized to
\(128\times128\) and 70k fake images of the same size for each StyleGAN model, i.e.,
StyleGAN, StyleGAN2, and StyleGAN3. As in \cite{wolter2021wavelet}, we split the full
dataset into training, validation, and test datasets with sizes of 252k, 8k, and 20k,
respectively. The CelebA and LSUN bedroom samples are generated following the recommendations
of the from the references \cite{frank2020leveraging, wolter2021wavelet},
which suggest using the pre-trained models from \cite{yu2019attributing}. Analogously to
\cite{wolter2021wavelet}, the fake dataset of FFHQ is instead generated using
the pre-trained models provided by the authors of
\cite{karras2019style,karras2020analyzing,karras2021alias}, using, in particular,
the R configuration (translation and rotation equivalent) of StyleGAN3,
a value of 1 for the truncation parameter, and image numbers from 1 to 70k as the random seeds. 

Secondly, the fully separable wavelet transform is realized by using the GPU enabled
fast wavelet transform library \verb+ptwt+ (\cite{pywt,WolterPhD,Blanke2021}). 
The samplets, implemented in C++, are integrated into the toolchain by using
\verb|pybind11| (\cite{pybind11}). The implementation is available at
\url{https://github.com/muchip/fmca}.

In order to demonstrate the advantages of features extracted by anisotropic multiresolution analyses,
we train the same shallow CNN proposed in \cite{frank2020leveraging},
fed with the samplets and the fully separable decomposition to match the state of the art performance.  
In the architecture, a fully connected layer is added at the end of the convolution part as a classifier.
Its output is the number of classes, and the input size is \(32d^2\), where \(d\) depends on the size
of input features. Because the samplet transform does not involve any padding procedure,
\(d\) for the samplets is the same number as for the raw pixels. However, the size of the fully separable decomposition coefficients depends on the padding methods, which is slightly larger than the size of
the raw pixels and the samplets except when using boundary filters. Therefore, the fully connected layer
for the FSWT without the boundary method is slightly larger than that for the raw pixels and samplets.
With a specific size for the Daubechies 3 orthogonal wavelet (\textit{db3}),
and reflect-padding method the total number of parameters for the FSWT increase to
roughly 202k compared to the 170k parameters for the raw pixels, DCT, and samplets.
Even though the numbers of parameters for samplets and FSWT are the same as for DCT and, thus,
larger than for wavelet packets,
their parameters are still only around \(2\%\) of the parameters in
\cite{yu2019attributing}. Besides, the training and evaluation procedures are faster using
samplets or FSWT compared to wavelet packets due to their linear time complexity.

In the training procedure, we set the batch size to 128 and train the model using the Adam algorithm (\cite{kingma2014adam}) with a learning rate of 0.001 for 10 epochs. The random seed values used
are 0,1,2,3,4 for 5 repetitive training procedures. The random seed is not only used to initialize
the network weights, but also shuffle the entire datasets before splitting.
This way, there is no bias in the partitioning of the data into the training,
test, and validation datasets.

\iffalse
\begin{table}[ht]
    \centering
    {\small
    \begin{tabularx}{0.5\linewidth}{cc}
        \toprule
        \multicolumn{2}{c}{Simple CNN} \\
        \hline
        Conv & (3,3,3,3)\\
        ReLU &\\
        Conv & (3,8,3,3) \\
        ReLU & \\
        AvgPool & (2,2) \\
        Conv & (8,16,3,3) \\
        ReLU & \\
        AvgPool & (2,2) \\
        Conv & (16,32,3,3) \\
        ReLU & \\ \hline
        Dense & (32 $\cdot$ d $\cdot$ d, $c$) \\
        \bottomrule
    \end{tabularx}
    }
    \caption{CNN classifier architecture for the samplets and the fully separable transform.}
    \label{tab:cnn_arch}
\end{table}
\fi

\begin{table*}
	\centering
	\begin{adjustbox}{width=\linewidth}
	\begin{tabular}{lccccccccccc}
		\toprule
		&& && && \multicolumn{2}{c}{CelebA\%} && \multicolumn{2}{c}{LSUN bedrooms\%}\\
		\cline{7-8} \cline{10-11}
		Method && complexity && parameters && max\ & $\mu\pm\sigma$ && max\ & $\mu\pm\sigma$\\ \hline
		Pixels (Frank \etal \cite{frank2020leveraging}) && $\mathrm{O}(1)$ && 170k && 97.80 & - && 98.95 & -\\
		DCT (Frank \etal \cite{frank2020leveraging}) && $\mathrm{O}(n^2\log n)$  && 170k && 99.07 & - && 99.64 & -\\
		Packet-ln-db3 (Wolter \etal \cite{wolter2021wavelet}) && $\mathrm{O}(n^2\log n)$  && 109k && 99.38 & $99.11\pm0.49$ && 99.19 & $99.01\pm0.17$\\
		Packet-ln-db4 (Wolter \etal \cite{wolter2021wavelet}) && $\mathrm{O}(n^2\log n)$  && 109k && 99.43 & $99.27\pm0.15$ && 99.02 & $98.46\pm0.67$\\ \hline
		Samplets-BN-2-3 (our) && $\mathrm{O}(n^2)$  && 170k && 99.84 & $99.71\pm0.12$ && 99.49 & $99.31\pm0.15$\\
		Samplets-BN-3-3 (our) && $\mathrm{O}(n^2)$&& 170k && 99.92 & $99.86\pm0.06$ && 99.44 & $99.36\pm0.09$\\
		Samplets-BN-4-3 (our) &&$\mathrm{O}(n^2)$ && 170k && 99.79 & $99.75\pm0.06$ && 99.2 & $98.96\pm0.21$\\
		Samplets-BN-5-3 (our) &&$\mathrm{O}(n^2)$ && 170k && 99.87 & $99.79\pm0.05$ && 99.52 & $99.34\pm0.11$\\
		Samplets-BN-6-3 (our) &&$\mathrm{O}(n^2)$ && 170k && 99.87 & $99.82\pm0.04$ && 99.55 & $99.42\pm0.13$\\
		Samplets-BN-7-3 (our) &&$\mathrm{O}(n^2)$ && 170k && 99.89 & $99.77\pm0.12$ && 99.62 & $99.38\pm0.24$\\
		FSWT-BN-db3-3-boundary (our) &&$\mathrm{O}(n^2)$ && 170k && 99.88 & $99.84\pm0.05$ && 99.1 & $98.76\pm0.35$\\
		FSWT-BN-db4-3-boundary (our) &&$\mathrm{O}(n^2)$ && 170k && 99.91 & $99.76\pm0.11$ && 99.33 & $99.13\pm0.18$\\
		FSWT-BN-db3-3-reflect (our) && $\mathrm{O}(n^2)$&& 202k && \textbf{99.97} & $99.96\pm0.01$ && \textbf{99.9} & $99.79\pm0.06$\\
		FSWT-BN-db4-3-reflect (our) &&$\mathrm{O}(n^2)$ && 225k && 99.96 & $99.86\pm0.14$ && 99.86 & $99.79\pm0.04$\\
		
		\bottomrule
	\end{tabular}
	\end{adjustbox}
	\caption{Accuracy results for different features with max, mean and standard deviation on the CelebA and the LSUN bedroom datasets. We report our results in comparison to the results in \cite{frank2020leveraging, wolter2021wavelet}, which are obtained using seeds from 0 to 4.}
	\label{tab:res_kfold_celeba_lsun}
\end{table*}

\begin{figure*}
	\centering
	\includegraphics[width=\linewidth]{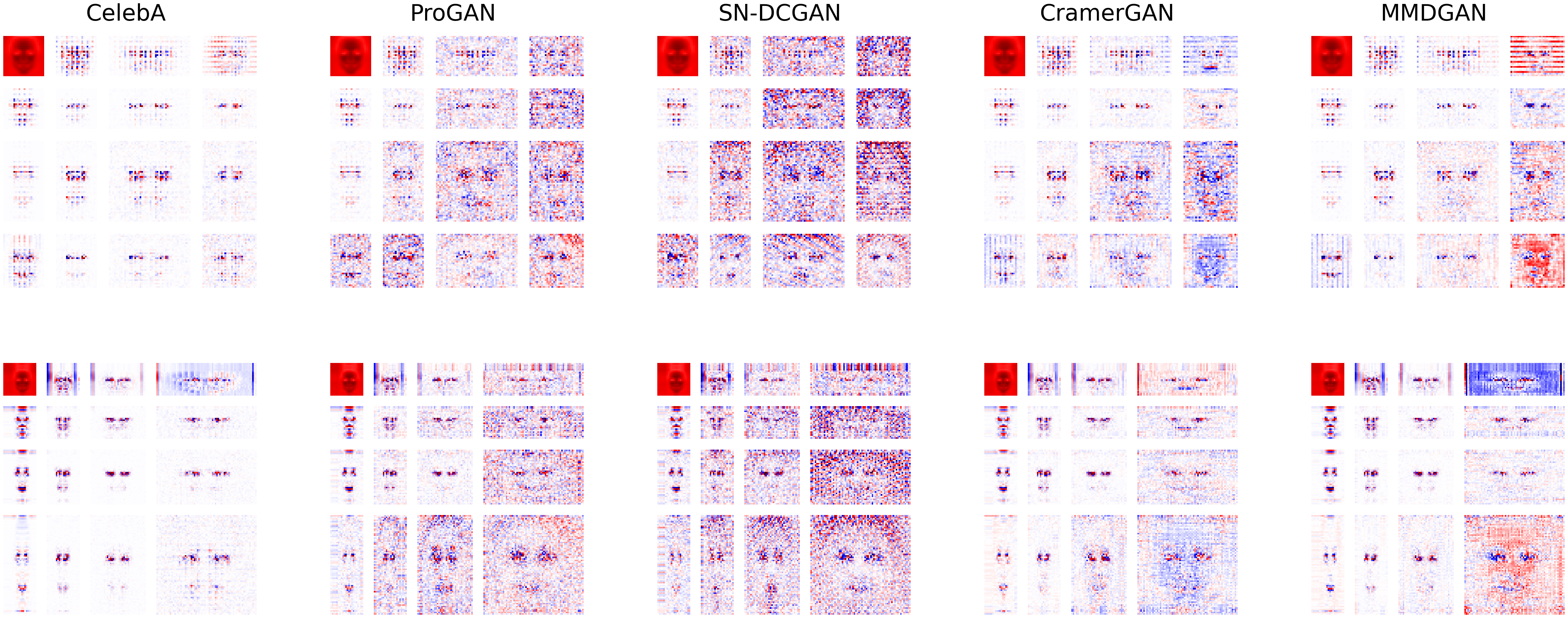}
	\caption{Average coefficients of each feature extraction method on the CelebA dataset.
	These visualizations are obtained by applying the block-wise normalization,
	then scaling them to the range \((0,255)\), and finally casting to unsigned int 8 bit.
	The first row corresponds Samplets-BN-3-3 and the second FSWT-BN-db3-3-reflect.
	We are using a diverging color map with white
	standing for zeros, red for positive numbers, and blue for negative numbers.}
	\label{fig:celeba_fs}
\end{figure*}

\noindent\textbf{Results:}
The best result from \cite{wolter2021wavelet} is obtained by the \textit{db4-fuse}
architecture, where a much more complex network than the one we use
is fed with both the wavelet packet representations, the Fourier transform,
and the raw pixels. We decided to ignore the results of the fuse architecture
because they rely more on the underlying network architecture than the input features.
We rather compare our features to the features in \cite{frank2020leveraging, wolter2021wavelet}
using the same simple CNN in \cite{frank2020leveraging}, 
focusing on the results solely obtained from a single multiresolution analysis:
wavelet packets \textit{db3} and \textit{db4}. We also evaluate samplets with 
vanishing moments up to order \(m=7\) to see how the accuracy changes when
increasing drastically the number of vanishing moments. Afterwards, we represent the block-wise
normalized samplets with vanishing moment up to order \(m\), and level of \(l\), denoted by
\emph{Samplets-BN-\(m\)-\(l\)}.

To summarize, for our features, we consider:
\begin{itemize}
    \item Samplets with vanishing moments up to order \(m=7\), and \(l=3\) as in \cite{wolter2021wavelet};
    \item Fully separable \textit{db3} and \textit{db4} with the maximum decomposition level of 3,
    to have a direct comparison to the wavelet packet \textit{db3} and \textit{db4};
\end{itemize}

From Table \ref{tab:res_kfold_celeba_lsun}, we see that on the datasets of CelebA and LSUN bedroom
the anisotropic transforms performs better in terms of the maximum, the mean, and the standard deviation.
Among all features, the fully separable \textit{db3} with the reflect-padding method is the clear winner.
On the another hand, if we only consider the fully separable transformation without padding, 
samplets with the vanishing moment 3 perform better than the FSWT with the boundary method on all aspects.
Samplets with a lot of vanishing moments do not bring much improvements compared
to the ones with lower vanishing moments. Thus, we will only focus on the samplets with vanishing moment up to the orders 3 and 4 to have a direct comparison to the the wavelet packet \textit{db3} and \textit{db4}
in our following experiments. To visually demonstrate why anisotropic features work better,
in Figure \ref{fig:celeba_fs}, we compare the average samplets, and the fully separable decomposition 
coefficients of the reference CelebA dataset and synthetic data generated from multiple GANs.
As we can see, the anisotropic analysis leads to different patterns for images from different
sources, which, we believe, could aid a neural classifier in discerning the origin of each data point.

We also want to explore how well the different features perform on more advanced GAN models.
Since the five GANs considered for the CelebA and the LSUN bedrooms are not state-of-the-art anymore,
we shift our focus to StyleGAN, StyleGAN2 and StyleGAN3, for which there also exist pre-trained
models on the FFHQ dataset. We repeat the classification task for 5 times and report the results in
Table \ref{tab:res_kfold_ffhq_celeba} for the six best methods identified before: Samplets-BN-3-3 and Samplets-BN-4-3,
fully separable \textit{db3} and \textit{db4} with the boundary and the reflect-padding methods, plus the samplets with a larger level \(l\), 
as well as the benchmarks in \cite{frank2020leveraging, wolter2021wavelet}. Even though samplets
perform not as good as on the CelebA and LSUN bedroom datasets, the fully separable \textit{db3}
and \textit{db3} with reflecting padding method consistently perform better compared
to the other features. 
\begin{table*}[ht]
    \centering
    \begin{adjustbox}{width=0.8\linewidth}
    \begin{tabular}{lccccc}
        \toprule
         & \multicolumn{2}{c}{FFHQ\%} && \multicolumn{2}{c}{incomplete CelebA\%}\\
        \cline{2-3} \cline{5-6}
        Method & max\ & $\mu\pm\sigma$ && max\ & $\mu\pm\sigma$\\\hline
        Pixels (Wolter \etal \cite{wolter2021wavelet}) & 93.71 & $90.90 \pm 2.19$ && 96.33 & - \\
        DCT (Frank \etal \cite{frank2020leveraging}) & - & -  && 98.47 & - \\
        Packet-ln-db4 (Wolter \etal \cite{wolter2021wavelet}) & 96.28 & $95.85 \pm 0.59$ && 99.01 & $96.96 \pm 3.47$ \\ \hline
        Samplets-BN-3-3 & 92.64 & $91.95\pm0.73$ && 99.53 & $99.3\pm0.12$\\
        Samplets-BN-3-4 & 94.45 & $92.6\pm1.03$ && 99.5 & $99.2\pm0.24$\\
        Samplets-BN-4-3 & 90.68 & $89.04\pm0.98$ && 98.82 & $98.14\pm0.42$\\
        Samplets-BN-4-4 & 90.24 & $88.52\pm1.05$ && 98.65 & $97.9\pm0.4$\\
        FSWT-BN-db3-3-boundary & 95.72 & $94.9\pm0.47$ && 99.18 & $98.81\pm0.24$\\
        FSWT-BN-db4-3-boundary & 94.35 & $93.63\pm0.7$ && 99.14 & $98.76\pm0.23$ \\
        FSWT-BN-db3-3-reflect & 96.36 & $95.13\pm1.63$ && 99.77 & $99.49\pm0.3$\\
        FSWT-BN-db4-3-reflect & \textbf{97.15} & $96.13\pm0.55$ && \textbf{99.89} & $99.68\pm0.16$ \\
        \bottomrule
    \end{tabular}
    \end{adjustbox}
    \caption{Accuracy results for different features with max, mean and standard deviation
    on the FFHQ dataset and a fifth of the CelebA data.
    We report the results as stated in \cite{wolter2021wavelet},
    which were obtained using seeds from 0 to 4.}
    \label{tab:res_kfold_ffhq_celeba}
\end{table*}

\subsection{Training on 20\% of the training dataset}
We also observe that anisotropic features can achieve a
higher accuracy than DCT and wavelet packets when not much training data is available.
To demonstrate this, we only train our model on 20\% of the CelebA training dataset,
and report the statistics of the accuracy on the entire CelebA test dataset. From Table \ref{tab:res_kfold_ffhq_celeba}, we observe that samplets and the FSWT performs better on all
aspects, especially on the average and the standard deviation. The fully separable
\textit{db4} with the reflect-padding method almost reaches
the same accuracy as using the entire training dataset.

\subsection{Robustness to Perturbations}
\begin{table*}[ht]
	\centering
	\begin{adjustbox}{width=0.8\linewidth}
	\begin{tabular}{lccccccc}
		\toprule
		& Blur\% & &Cropped\%& & Compression\%& & Noise\%\\ \hline
		Pixels (Frank \etal \cite{frank2020leveraging}) & 88.23 & & 97.82 & & 78.67 & & 78.18\\
		DCT (Frank \etal \cite{frank2020leveraging}) & 93.61 & & 98.83 & & 94.83 & & 89.56\\
		Packet-ln-db3 (Wolter \etal \cite{wolter2021wavelet}) & - & & 95.68 & & 84.73 & & -\\ \hline
		Samplets-BN-3-3 & 94.29 & &98.15 & & 92.51 & & 86.45\\
		Samplets-BN-4-3 & 93.1 & &97.87 & & 90.55 & & 83.46\\
		FSWT-BN-db3-3-boundary & 91.14 & & 98.49 & & 94.23 & & 88.17\\
		FSWT-BN-db4-3-boundary & 90.71 & & 98.35 & & 95.19 & & 86.41\\
		FSWT-BN-db3-3-reflect & \textbf{96.79} & & \textbf{99.2} & & 96.86 & & \textbf{90.88} \\
		FSWT-BN-db4-3-reflect & 96.28 & & 99.13 & & \textbf{97.23} & & 90.8\\
		\bottomrule
	\end{tabular}
	\end{adjustbox}
	\caption{Accuracy results for different features with max on the distorted datasets}
	\label{tab:perb_lsun}
\end{table*}
Finally, we test the resilience of different features against the 4 common image perturbations.
We consider the same image perturbation configurations as in \cite{frank2020leveraging}:
Gaussian blurring with a kernel size randomly sampled from \((3, 5, 7, 9)\),
image crop by a percentage randomly selected from \(U(5,20)\),
JPEG based compression with a quality randomly selected from \(U(10, 75)\),
and addition of Gaussian noise whose variance is sampled from the uniform distribution \(U(5,20)\). The modified pixel values are clipped to the range of (0,255), followed by being cast into 8-bit unsigned integers. In the experiment, we apply one of the mentioned perturbations with a probability
of \(\frac{1}{2}\). Here, we conduct the experiments on the LSUN bedroom dataset. Results
are shown in Table \ref{tab:perb_lsun}. Most anisotropic transformations perform much better
than the other features including pixels, DCT coefficients, and wavelet packets.
Our features are very robust when images are exposed to the random crop. Instead
the accuracy is reduced more dramatically under the blurring, JPEG-based compression,
and noise addition. Especially in the case of noise addition, the accuracy is reduced by around 10\%.

\section{Conclusion}
With the results for anisotropic multiresolution analyses presented,
we have added a new aspect to the general understanding
of how GANs truly operate and what traces they tend to leave in their output.
Based on the experiments, the fully separable decomposition and samplets improved
the accuracy results compared to the state-of-the-art on the CelebA and LSUN bedroom datasets.
However, even though samplets and FSWT with the boundary method do not achieve the state-of the art
on the FFHQ dataset, FSWT with the reflect-padding method perform consistently best among all features.
In terms of training on incomplete datasets and robustness to perturbations,
anisotropic transformations are better than wavelet packets. Moreover, samplets and FSWT
with the boundary method do not require padding. They are thus free from boundary artifacts,
allowing them to transform any input signal while maintaining the same support size.
Furthermore, this capability makes them perfect as a drop-in addition to any network architecture,
which means that, for example, any pixel-based discriminator could instantly and effortlessly
improve its classification performance by just adding a single preprocessing layer without
modifying its architecture.

%%%%%%%%% REFERENCES
\bibliography{egbib}

\end{document}